\documentclass[sigconf]{acmart}

\AtBeginDocument{%
  \providecommand\BibTeX{{%
    \normalfont B\kern-0.5em{\scshape i\kern-0.25em b}\kern-0.8em\TeX}}}

\copyrightyear{2025}
\acmYear{2025}
\setcopyright{rightsretained}
\acmConference[CHI EA '25]{Extended Abstracts of the CHI Conference on Human Factors in Computing Systems}{April 26-May 1, 2025}{Yokohama, Japan}
\acmBooktitle{Extended Abstracts of the CHI Conference on Human Factors in Computing Systems (CHI EA '25), April 26-May 1, 2025, Yokohama, Japan}\acmDOI{10.1145/3706599.3706690}
\acmISBN{979-8-4007-1395-8/25/04}

\usepackage{hyperref}
\usepackage{enumitem}
\usepackage{xspace}
\usepackage{xcolor}
\usepackage[capitalise, noabbrev]{cleveref}
\usepackage{subcaption}
\usepackage{multirow}
\usepackage{colortbl}
\usepackage{listings}
\usepackage{wrapfig}

\newcommand{\eg}{\emph{e.g.,}\xspace}%

\newcommand{\pf}{\emph{Find-Fix-Verify}\xspace}
\newcommand{\pRecrusive}{\emph{Price-Divide-Solve}\xspace}
\newcommand{\pMapReduce}{\emph{Map-Reduce}\xspace}
\newcommand{\pHumor}{\emph{HumorTool}\xspace}
\newcommand{\pIterateParallel}{\emph{Iterative Process}\xspace}
\newcommand{\pIterative}{\emph{Microtasking}\xspace}
\newcommand{\pSRL}{\emph{Task Paraphrase}\xspace}
\newcommand{\opensourceurl}{\url{https://github.com/tongshuangwu/llm-crowdsourcing-pipeline/}}

\definecolor{clightgrey}{HTML}{80868b}
\definecolor{cgenerate}{HTML}{3B4D73}
\definecolor{cprompt}{HTML}{3c4043}
\definecolor{cexample}{rgb}{0.23, 0.30, 0.45}

\newcommand{\quoteinline}[1]{{\color{cprompt}\emph{``#1''}}\xspace}

\newcolumntype{L}[1]{>{\raggedright\let\newline\\\arraybackslash\hspace{0pt}}m{#1}}
\newcolumntype{C}[1]{>{\centering\let\newline\\\arraybackslash\hspace{0pt}}m{#1}}
\newcolumntype{R}[1]{>{\raggedleft\let\newline\\\arraybackslash\hspace{0pt}}m{#1}}

\newcommand{\paragraphBold}[1]{\paragraph{\emph{\textbf{#1}}}}

\begin{document}

\title[Replicating Crowdsourcing Pipelines with LLMs]{LLMs as Workers in Human-Computational Algorithms? \\
Replicating Crowdsourcing Pipelines with LLMs}


\author{Tongshuang Wu}
\orcid{0000-0003-1630-0588}
\email{sherryw@cs.cmu.edu}
\affiliation{%
  \institution{Carnegie Mellon University}
  \city{Pittsburgh}
  \state{PA}
  \country{USA}
  \postcode{15213}
}
\thanks{\ \ This work builds upon an assignment from CMU 05-499/899: Human-Centered NLP. Tongshuang Wu is the course instructor, who designed the assignment based on discussions with Haiyi Zhu, and wrote the majority of the paper. The remaining authors are students listed in alphabetical order. Following the principles of participatory research, all students were informed of the research, chose to participate, and were given opportunities to cease participation/authorship after the initial experiments and after reviewing the paper draft.} 

\author{Haiyi Zhu}
\orcid{0000-0001-7271-9100}
\email{haiyiz@cs.cmu.edu}
\affiliation{%
  \institution{Carnegie Mellon University}
  \city{Pittsburgh}
  \state{PA}
  \country{USA}
  \postcode{15213}
}

\author{Maya Albayrak}
\orcid{0009-0006-7684-4981}
\email{malbayra@alumni.cmu.edu}
\affiliation{%
  \institution{Carnegie Mellon University}
  \city{Pittsburgh}
  \state{PA}
  \country{USA}
  \postcode{15213}
}

\author{Alexis Axon}
\orcid{0009-0003-0445-1810}
\email{aaxon@alumni.cmu.edu}
\affiliation{%
  \institution{Carnegie Mellon University}
  \city{Pittsburgh}
  \state{PA}
  \country{USA}
  \postcode{15213}
}

\author{Amanda Bertsch} 
\orcid{0000-0002-1368-1111}
\email{abertsch@andrew.cmu.edu}
\affiliation{%
  \institution{Carnegie Mellon University}
  \city{Pittsburgh}
  \state{PA}
  \country{USA}
  \postcode{15213}
}

\author{Wenxing Deng} 
\orcid{0009-0007-0380-8397}
\email{wenxing_deng@outlook.com}
\affiliation{%
  \institution{Carnegie Mellon University}
  \city{Pittsburgh}
  \state{PA}
  \country{USA}
  \postcode{15213}
}

\author{Ziqi Ding}
\orcid{0009-0004-5464-4546}
\email{ziqiding@alumni.cmu.edu}
\affiliation{%
  \institution{Carnegie Mellon University}
  \city{Pittsburgh}
  \state{PA}
  \country{USA}
  \postcode{15213}
}

\author{Boyuan Guo}
\orcid{0000-0002-8303-4649}
\email{boyuang@andrew.cmu.edu}
\affiliation{%
  \institution{Carnegie Mellon University}
  \city{Pittsburgh}
  \state{PA}
  \country{USA}
  \postcode{15213}
}

\author{Sireesh Gururaja}
\orcid{0009-0004-0309-3393}
\email{sgururaj@andrew.cmu.edu}
\affiliation{%
  \institution{Carnegie Mellon University}
  \city{Pittsburgh}
  \state{PA}
  \country{USA}
  \postcode{15213}
}

\author{Tzu-Sheng Kuo}
\orcid{0000-0002-1504-7640}
\email{tzushenk@cs.cmu.edu}
\affiliation{%
  \institution{Carnegie Mellon University}
  \city{Pittsburgh}
  \state{PA}
  \country{USA}
  \postcode{15213}
}

\author{Jenny T. Liang}
\orcid{0000-0001-6722-9959}
\email{jtliang@cs.cmu.edu}
\affiliation{%
  \institution{Carnegie Mellon University}
  \city{Pittsburgh}
  \state{PA}
  \country{USA}
  \postcode{15213}
}

\author{Ryan Liu}\email{rl5886@princeton.edu}
\orcid{0009-0003-8755-649X}
\affiliation{%
  \institution{Carnegie Mellon University}
  \city{Pittsburgh}
  \state{PA}
  \country{USA}
  \postcode{15213}
}

\author{Ihita Mandal}
\orcid{0009-0001-0705-8390}
\email{imandal@andrew.cmu.edu}
\affiliation{%
  \institution{Carnegie Mellon University}
  \city{Pittsburgh}
  \state{PA}
  \country{USA}
  \postcode{15213}
}

\author{Jeremiah Milbauer}
\orcid{0000-0003-2809-534X}
\email{jmilbaue@cs.cmu.edu}
\affiliation{%
  \institution{Carnegie Mellon University}
  \city{Pittsburgh}
  \state{PA}
  \country{USA}
  \postcode{15213}
}

\author{Xiaolin Ni}
\orcid{0009-0009-2660-9539}
\email{xiaolinjy@gmail.com}
\affiliation{%
  \institution{Carnegie Mellon University}
  \city{Pittsburgh}
  \state{PA}
  \country{USA}
  \postcode{15213}
}

\author{Namrata Padmanabhan}
\orcid{0009-0008-1617-9566}
\email{npadmana@andrew.cmu.edu}
\affiliation{%
  \institution{Carnegie Mellon University}
  \city{Pittsburgh}
  \state{PA}
  \country{USA}
  \postcode{15213}
}

\author{Subhashini Ramkumar}
\orcid{0009-0005-6448-562X}
\email{askforsubhashini@gmail.com}
\affiliation{%
  \institution{Carnegie Mellon University}
  \city{Pittsburgh}
  \state{PA}
  \country{USA}
  \postcode{15213}
}

\author{Alexis Sudjianto}
\orcid{0009-0003-9946-9422}
\email{asudjian@andrew.cmu.edu}
\affiliation{%
  \institution{Carnegie Mellon University}
  \city{Pittsburgh}
  \state{PA}
  \country{USA}
  \postcode{15213}
}

\author{Jordan Taylor}
\orcid{0000-0002-0896-992X}
\email{jordant@andrew.cmu.edu}
\affiliation{%
  \institution{Carnegie Mellon University}
  \city{Pittsburgh}
  \state{PA}
  \country{USA}
  \postcode{15213}
}

\author{Ying-Jui Tseng}
\orcid{0009-0006-1801-6061}
\email{yingjuit@alumni.cmu.edu}
\affiliation{%
  \institution{Carnegie Mellon University}
  \city{Pittsburgh}
  \state{PA}
  \country{USA}
  \postcode{15213}
}

\author{Patricia Vaidos}
\orcid{0009-0002-4189-8307}
\email{pvaidos@alumni.cmu.edu}
\affiliation{%
  \institution{Carnegie Mellon University}
  \city{Pittsburgh}
  \state{PA}
  \country{USA}
  \postcode{15213}
}

\author{Zhijin Wu}
\orcid{0009-0009-1875-5827}
\email{zhijinw@andrew.cmu.edu}
\affiliation{%
  \institution{Carnegie Mellon University}
  \city{Pittsburgh}
  \state{PA}
  \country{USA}
  \postcode{15213}
}

\author{Wei Wu}
\orcid{0009-0008-1671-9470}
\email{weiwu3@alumni.cmu.edu}
\affiliation{%
  \institution{Carnegie Mellon University}
  \city{Pittsburgh}
  \state{PA}
  \country{USA}
  \postcode{15213}
}

\author{Chenyang Yang}
\orcid{0000-0001-5016-7296}
\email{cyang3@andrew.cmu.edu}
\affiliation{%
  \institution{Carnegie Mellon University}
  \city{Pittsburgh}
  \state{PA}
  \country{USA}
  \postcode{15213}
}

\renewcommand{\shortauthors}{Wu, et al.}

\begin{abstract}

LLMs have shown promise in replicating human-like behavior in crowdsourcing tasks that were previously thought to be exclusive to human abilities. However, current efforts focus mainly on simple atomic tasks.
We explore whether LLMs can replicate more complex crowdsourcing pipelines. We find that modern LLMs can simulate some of crowdworkers' abilities in these ``human computation algorithms,'' but the level of success is variable and influenced by requesters' understanding of LLM capabilities, the specific skills required for sub-tasks, and the optimal interaction modality for performing these sub-tasks. 
We reflect on human and LLMs' different sensitivities to instructions, stress the importance of enabling human-facing safeguards for LLMs, and discuss the potential of training humans and LLMs with complementary skill sets.
Crucially, we show that replicating crowdsourcing pipelines offers a valuable platform to investigate 1) the relative LLM strengths on different tasks (by cross-comparing their performances on sub-tasks) and 2) LLMs' potential in complex tasks, where they can complete part of the tasks while leaving others to humans.
\end{abstract}




\maketitle

\section{Introduction}
\label{sec:intro}

The rapid advancement of AI systems has revolutionized our understanding of the capabilities of machines. 
One remarkable breakthrough in this field is the emergence of LLMs like ChatGPT. 
With a combination of extensive pre-training~\cite{brown2020language} and instruction tuning~\cite{stiennon2020learning, wu2023fine}, LLMs now not only possess a large amount of world knowledge, but can effectively leverage this knowledge to accomplish various tasks simply by following instructions.

Various studies have reported that these models can replicate human-like behavior to some extent, which is a key objective in the training of AI models~\cite{wang2021measure, bubeck2023sparks}. 
In particular, many studies focus on using LLMs to replicate crowdsourcing tasks, as they represent a wide range of tasks that were previously considered exclusive to human computational capabilities~\cite{bernstein2013crowd}.
For example, LLMs can generate annotations of higher quality at a reduced cost compared to crowdworkers or even experts~\citep{gilardi2023chatgpt, tornberg2023chatgpt}, and can approximate human opinions in subjective tasks, allowing for simulated human responses to crowdsourced questionnaires and interviews~\citep{hamalainen2023evaluating, argyle2022out}.
These observations indicate that LLMs will have significant social and economic implications, potentially reshaping the workforce by replacing certain human jobs~\citep{eloundou2023gpts}. 
In fact, crowdworkers now tend to rely on LLMs for completing text production tasks~\cite{veselovsky2023artificial}.

\begin{figure}[t]
\centering
\includegraphics[trim={0 27cm 36cm 0cm}, clip, width=0.83\linewidth]{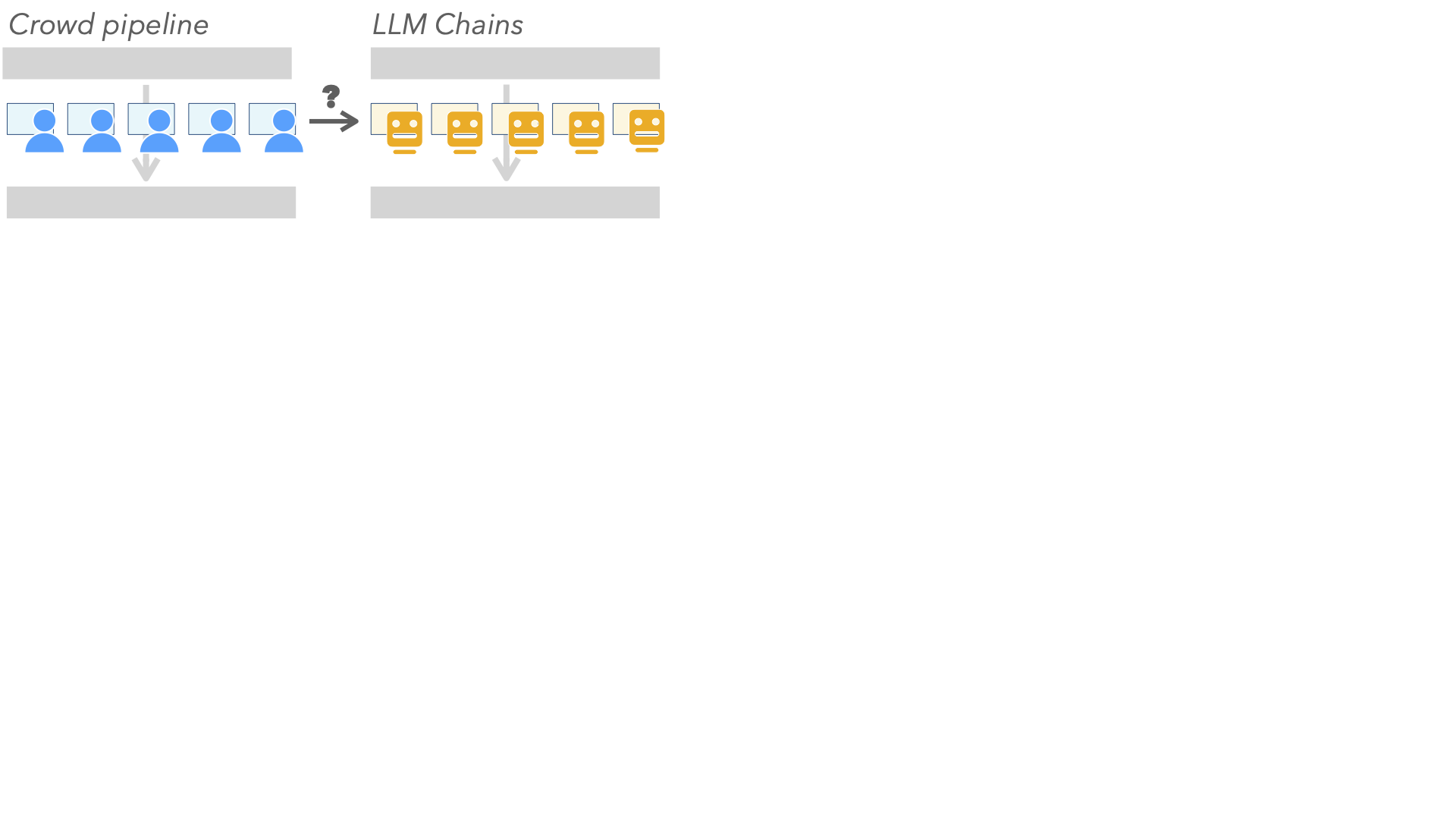}
\vspace{-10pt}
\caption{
We study whether LLMs can be used to replicate crowdsourcing pipelines and replace human workers in certain advanced ``human-computational process.''
}
\Description{An illustrative figure illustrating the main paper idea: whether LLMs can be used to replicate crowdsourcing pipelines and replace human workers in certain advanced ``human-computational process.''}
\label{fig:teaser}
\end{figure}

However, most existing efforts tend to focus on atomic tasks that are simple, self-contained, and easy for a single crowdworker to complete in a short amount of time --- the most basic version of \emph{human computational power}.
These efforts also are scattered across various tasks and domains, making it hard to systematically compare and understand which tasks LLMs may excel or underperform at, and to what extent they can simulate, replace, or augment humans on specific tasks.
Such emphases prompt us to ask, \emph{how far does the LLM replicability generalize?} Will they be useful in \emph{more advanced formats} of ``human computation''?

We are especially interested in whether LLMs can be used to replicate \emph{crowdsourcing pipelines}, which represent a more sophisticated approach to harnessing human computation~\cite{little2010exploring}. 
In a typical pipeline, complex tasks are broken down into pieces (\emph{sub-tasks}) that can be performed independently, then later combined~\cite{chilton2013cascade, kim2017mechanical, law2011towards, retelny2017no}.
This method has been widely used to scale crowdsourcing usability, allowing it to handle tasks that are too challenging for individual crowdworkers with limited level of commitment and unknown expertise~\cite[e.g., summarizing lengthy novels, software development, or deciphering heavily blurred text;][]{kittur2011crowdforge}.

Interestingly, research on LLMs has also explored scaling their capabilities for more complex tasks through \emph{chaining}.
Though named differently, LLM chains and crowdsourcing pipelines share similar motivation and strategy. 
Previous studies have connected the two, noting that they decompose tasks to address different problems~\cite{wu2022ai}: crowdsourcing pipelines focus on factors affecting human worker performance, such as cognitive load and task duration, while LLM chains address inherent limitations of LLMs, such as high variance in prompt effectiveness.
However, since LLMs have now been trained to better align with humans in following instructions and handling complex contexts~\cite{ouyang2022training}, it is possible for human and LLM workers to adopt the same task division strategies~\cite{grunde2023designing}. 

In this study, we investigate the potential of LLMs to replace human workers in advanced human computation processes. 
To do so, we designed an assignment for 
special topic course named \emph{Human-Centered NLP} at Carnegie Mellon University. 
In the assignment, 20 students were tasked to select one (out of seven) crowdsourcing pipelines depicted in prior work, and replicate them by employing LLMs to handle each sub-task.
The replication study also offers an interesting bonus analysis point: While LLM modules in a chain perform unique sub-tasks, all the sub-tasks occur in the same application domain (e.g., processing the same document in different ways), making it fairer to compare LLMs' performance across sub-tasks and uncovering the relative strengths and weaknesses.

We find that while LMs appear to be able to replicate crowdsourcing pipelines, there is a wide variance in which parts they tend to perform well / in ways we would expect from humans (main findings in Table~\ref{table:summary_findings}).
The differences emerge from two primary reasons. 
First, \emph{LLMs and humans respond differently to instructions}. 
LLMs are more responsive to adjectives and comparison-based instructions, such as ``better'' or ``more diverse,'' whereas humans handle instructions involving trade-off criteria better.
Second, \emph{humans receive more scaffolds through disagreement resolution mechanisms and interface-enforced interactions}, enabling guardrails on output quality and structure that not available to LLMs.
These observations highlight the need to improve LLM instruction tuning to better handle ambiguous or incomplete instructions, as well as the necessity to consider how non-textual ``instructions'' can be employed either during LLM finetuning or actual usage.
Moreover, the effectiveness of replicated LLM chains depends on students' perceptions of LLM strengths, which calls for more investigations on assisted prompting. 

In addition to offering immediate insights into the differences between LLMs and crowdworkers, we demonstrates that replicating crowdsourcing pipelines serves as a valuable platform for future investigations into the \emph{partial effectiveness} of LLMs across a \emph{wider range of tasks}.
Rather than expecting LLMs to tackle entire complex tasks, we can instead identify specific sub-tasks in which LLMs consistently perform on par with humans. This evidence can then be utilized to distribute sub-tasks between LLMs and human workers, optimizing the allocation of responsibilities.
We opensource the 
prompt chains and evaluation at \opensourceurl.

\section{Study Design}

\textbf{Procedure.} 
The study required participants (students) to replicate a crowdsourcing pipeline by writing multiple prompts that instruct LLMs to complete different microtasks.

They began by thoroughly reading a crowdsourcing pipeline paper for replication; then to demonstrate the effectiveness of their replicated pipeline, they were also asked to determine an appropriate testing task, create at least three test cases with pairs of inputs and ideal outputs, and self-propose a set of task-dependent metrics for evaluating pipeline outputs (e.g., fluency, creativity, coherence).
Then, they were instructed to implement two solutions: (1) a baseline solution that prompts one LLM module to complete the entire task (\emph{Baseline}), and (2) a replica of their chosen crowdsourcing pipeline (\emph{LLM Chain}).
They compared the two LLM solutions using their designated test cases and metrics, providing the reasoning behind their ratings.
Finally, they concluded the task by reflecting on why the LLM chain replication either succeeded or failed and brainstormed possible ways to improve the chains in the future.

Submissions were then peer-reviewed by three classmates in a double-blind process. 
The peers rated the submissions based on replication correctness, thoroughness, and comprehensiveness of their envisioned LLM chain improvements.
They rated all the criteria on a five-level Likert Scale and supplied detailed reasoning for their grading. The instructor carefully reviewed the gradings and excluded any assessments that appeared to lack thoughtful reflections or misunderstood the submissions.
The assignment details and the student submissions and gradings are available at \opensourceurl.

\textbf{Participants.}
21 students (13 females, 8 males) completed the task as one of their assignments for a graduate level 
course 05-499/899: Human-Centered NLP.\footnote{\url{http://www.cs.cmu.edu/~sherryw/courses/2023s-hcnlp.html}}
This comprised of 6 undergraduates, 10 master's students, and 5 PhD students specializing in Sociology, Learning Science, HCI, or NLP. 
The paper presents findings from 20 students' submissions, as one student opted for a non-programming approach for partial credit.

\textbf{Crowdsourcing papers.}
We selected crowdsourcing papers based on three criteria: 
(1) \emph{Diversity}: the papers should cover different pipeline designs (iterative, parallel), intermediate steps (question-answering, comparison, editing), and tasks (creative tasks, annotation tasks, editing tasks, etc.)
(2) \emph{Replicability}: The papers should detail every sub-step and provide concrete examples. Considering our emphasis on LLMs, we exclusively considered papers that described tasks with textual inputs and outputs.
(3) \emph{Asynchronized}: For the ease of setup, the papers should allow (LLM) workers to complete their microtasks independently without synchronized discussions.
The instructor pre-selected six papers meeting these criteria (the first six in Table~\ref{tab:pipelines}), and students could propose additional papers for approval (\pSRL in Table~\ref{tab:pipelines}). 
Up to four students could sign up to replicate the same pipeline in a first-come-first-serve manner.

\renewcommand{\arraystretch}{1}
\begin{table*}[!ht]
\centering

\scalebox{0.79}{

\setlength{\tabcolsep}{3pt}
\vspace{-20pt}
\begin{tabular}{@{} 
    R{0.18\textwidth} | 
    L{0.43\textwidth} | 
    L{0.11\textwidth}  | 
    R{0.07\textwidth} | R{0.07\textwidth} | R{0.07\textwidth} | R{0.07\textwidth} @{} }
\toprule
\multirow{2}{*}{\textbf{Pipeline}}
& \multirow{2}{*}{\textbf{Description}}
& \multirow{2}{*}{\textbf{Sample Task}}
& \multicolumn{4}{c}{\textbf{Replication evaluation}} \\

& & & {Total} & {Unique} & {Correct} & {Effective} \\
\midrule\midrule

\textbf{\pMapReduce} \citep{kittur2011crowdforge} 
    & \emph{Partition} tasks into discrete subtasks, \emph{Map} subtasks to workers, \emph{Reduce} / merge their results into a single output
    & Write essay
    & 4 & 1 & 3 & 3\\
\arrayrulecolor{black!30}\midrule
\multicolumn{7}{l}{\includegraphics[trim={0 42.5cm 0cm 12.5cm}, clip,width=1\linewidth]{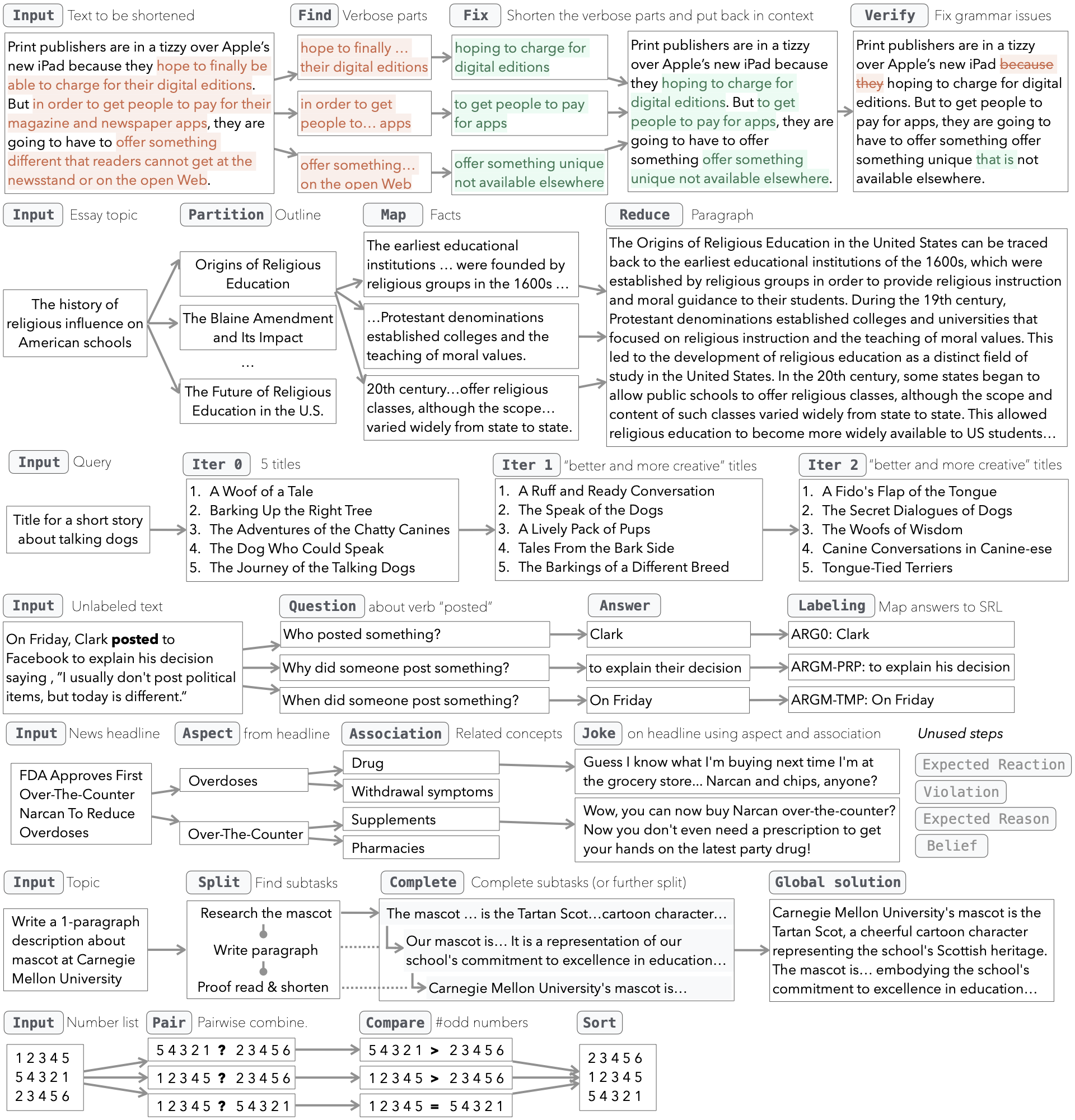}}\\
\arrayrulecolor{black!100}\midrule\midrule

\textbf{\pHumor} \citep{chilton2016humortools} 
    & Define semantic roles as the answers to a series of questions that are intuitive for non-experts.
    & Create satire 
    & 4 & 2 & 3 & 1  \\
\arrayrulecolor{black!30}\midrule
\multicolumn{7}{l}{\includegraphics[trim={0 16.3cm 0cm 45.5cm}, clip,width=1\linewidth]{figures/pipelines.png}}\\
\arrayrulecolor{black!100}\midrule\midrule

\textbf{\pIterateParallel} \citep{little2010exploring} 
    & Feed the result of one creation task into the next, so workers see content generated by previous workers. 
    & Brainstorm
    & 3 & 2 & 3 & 2 \\
\arrayrulecolor{black!30}\midrule
\multicolumn{7}{l}{\includegraphics[trim={0 33.9cm 0cm 28.4cm}, clip,width=1\linewidth]{figures/pipelines.png}}\\
\arrayrulecolor{black!100}\midrule\midrule

\textbf{\pIterative} \citep{cheng2015break}
    & Concrete microtasking for sorting task: an implementation of human-powered quicksort
    & Sorting
    & 3 & 3 & 3 & 1 \\
\arrayrulecolor{black!30}\midrule
\multicolumn{7}{l}{\includegraphics[trim={0 0cm 0cm 63.5cm}, clip,width=1\linewidth]{figures/pipelines.png}}\\
\arrayrulecolor{black!100}\midrule\midrule

\textbf{\pf} \citep{bernstein2010soylent} 
    &For writing and editing: \emph{Find} problems, \emph{Fix} the identified problems, \emph{Verify} these edits
    & Shorten text
    & 3 & 3 & 2 & 1 \\
\arrayrulecolor{black!30}\midrule
\multicolumn{7}{l}{\includegraphics[trim={0 58cm 0cm 0cm}, clip,width=1\linewidth]{figures/pipelines.png}}\\
\arrayrulecolor{black!100}\midrule\midrule

\textbf{\pRecrusive} \citep{kulkarni2012collaboratively} 
    & Workers recursively divide complex steps until they are at an appropriately simple level, then solve them.
    & Write essay
    & 1 & 1 & 1 & 1 \\
\arrayrulecolor{black!30}\midrule
\multicolumn{7}{l}{\includegraphics[trim={0 7.5cm 0cm 54.5cm}, clip,width=1\linewidth]{figures/pipelines.png}}\\
\arrayrulecolor{black!100}\midrule\midrule

\textbf{\pSRL} \citep{he2015question} 
    & Define semantic roles as the answers to a series of questions that are intuitive for non-experts.
    & SRL labeling 
    & 1 & 1 & 1 & 1 \\
\arrayrulecolor{black!30}\midrule
\multicolumn{7}{l}{\includegraphics[trim={0 25.5cm 0cm 37.4cm}, clip,width=1\linewidth]{figures/pipelines.png}}\\


\arrayrulecolor{black!100}\bottomrule
\end{tabular}
}
\caption{Crowdsourcing pipelines replicated, and their example outputs from student-replicated LLM chains.}
\Description{A table that shows that concrete seven crowdsourcing pipelines, and their example outputs from student-replicated LLM chains.}
\label{tab:pipelines}
\vspace{-30pt}

\end{table*}

\textbf{LLMs.} Students were required to use OpenAI \texttt{text-davinci-003}
for final implementations, the most capable model available at the time of assignment design.
However, they were encouraged to start with less expensive models like \texttt{text-ada-001} for initial exploration.

\textbf{Assessment.}
We evaluate the  chains on two dimensions: 
(1) \emph{Replication correctness}: We measure the success of replication using the peer grading results. A replication is considered successful if the average peer score for \emph{Correct Replication} is greater than three;
(2) \emph{Chain effectiveness}: We evaluate whether the replicated chains are more effective than the baselines using the students' own assessment. If students indicate that their replicated chains outperform the baselines on the majority of their tested inputs,the chain is considered effective.

Since multiple students replicated the same pipelines, we also compare replicas for the same pipeline to reveal key factors for successful replication.
To do so, we report the number of (3) \emph{Unique replicas}: We manually grouped the students' LLM chains based on the microtasks involved, deeming two chains identical if they include steps that essentially serve the same intended functionality, even if there are wording differences in the LLM prompts. 

\newcommand{\crowdpipe}{\textcolor{blue}{Crowd. pipelines}:\xspace}
\newcommand{\llmchain}{\textcolor{orange}{LLM chains}:\xspace}
\newcommand{\crowd}{\textcolor{blue}{Crowds}:\xspace}
\newcommand{\llm}{\textcolor{orange}{LLMs}:\xspace}
\newcommand{\shared}{\textcolor{gray}{Both}:\xspace}

\renewcommand{\arraystretch}{0.9}

\begin{table*}[ht]

\fontsize{8.5}{9}\selectfont
\centering
\setlength{\tabcolsep}{3pt}

\begin{subtable}[ht]{ 1\textwidth}
\centering

\scalebox{0.95}{
\begin{tabular}{@{} c | R{0.12\linewidth} | L{0.83\linewidth}  @{}}
\toprule
\multicolumn{2}{c|}{Dimensions} & Observations  \\

\midrule\midrule
\multirow{4}{*}{\rotatebox[origin=l]{90}{Pipelines}} 
& Idea
    & \shared Breakdown complex tasks into pieces that can be done independently, then combined. \\ 
\cmidrule(){2-3}
& Limitations
    & \shared Cascading errors, conflicts between parallel paths, etc. \\
\cmidrule(){2-3}
& Gains
    & 
    \shared Scale to tasks that are otherwise hard, more structured interactions, more resilient to interruptions.\newline
    \llmchain Can take advantage of cascading effects \& parallel paths, for explainability. \\
\cmidrule(){2-3}
& Optimal design
    & 
    \crowdpipe Address pitfalls of a single worker: high task variance, limited cognitive load, etc. \newline
    \llmchain Address pitfalls of a single LLM: limited reasoning capabilities, etc. \\
\bottomrule
\end{tabular}
}
\label{table:summary_prior}
\caption{Similarities between crowdsourcing pipelines and LLM chains summarized in prior work \cite[e.g.,][]{wu2022ai}.}
\Description{Similarities between crowdsourcing pipelines and LLM chains summarized in prior work}
\end{subtable}
\vspace{5pt}

\begin{subtable}[ht]{ 1\textwidth}
\centering
\scalebox{0.9}{
\begin{tabular}{@{} c | R{0.12\linewidth} | L{0.41\linewidth} | L{0.45\linewidth}  @{}}
\toprule
\multicolumn{2}{c|}{Dimensions} & Observations & Reflections \& Opportunities \\

\midrule\midrule
\rotatebox[origin=c]{90}{Pipelines}
& Practical design
    & \shared Can benefit from similar pipeline designs (as LLMs are finetuned on instructions). \newline
    \llmchain Vary based on students' beliefs about LLM strengths and weaknesses.
    & Develop frameworks that can enable practitioners to adapt their perception of LLM usefulness by adjusting prompt granularity.
    \\

\midrule\midrule
\multirow{2}{*}{\rotatebox[origin=l]{90}{Per-step / task}} 
& Sensitivity to instructions 
    & \crowd Can subconsciously balance trade-offs in instructions, vs. LLMs need explicit prioritization. \newline
      \llm Responsive to abstract instructions (``more diverse titles''), vs. crowdworkers face anchoring bias.
    & Assess the effects of LLM instruction tuning (e.g., sensitivity to adjectives,  emphasis on singular needs); \newline
    Tune LLMs to follow more ambiguous instructions; \newline 
    Train humans to develop skills complementary to LLM strengths.\\
    
\cmidrule(){2-4}
& Output quality scaffolds 
    & \crowd Noise and disagreement resolution  \newline
      \llm None; LLM non-determinism is overlooked.
    & Treat different LLM generations using the same prompt as votes of multiple LLM workers. \\
\cmidrule(){2-4}
& Output structure scaffolds 
    & \crowd Multimodal ``instructions'' (e.g., textual descriptions, interface regulations). \newline
    \llm Textual instructions only.
    & 
    Extend the human-LLM {alignment} to also consider optimal modality of instruction; \newline
    Map observations on LLM-simulated humans to actual humans.
    \\

\bottomrule
\end{tabular}
}
\caption{An overview of observations and reflections on students' replications on crowdsourcing pipelines.}

\label{table:summary_findings}

\end{subtable}
\vspace{-10pt}
\end{table*}

\section{Results and Reflection}

\begin{figure*}[!h]
\centering
\includegraphics[trim={0 8cm 16cm 0cm}, clip, width=0.9\linewidth]{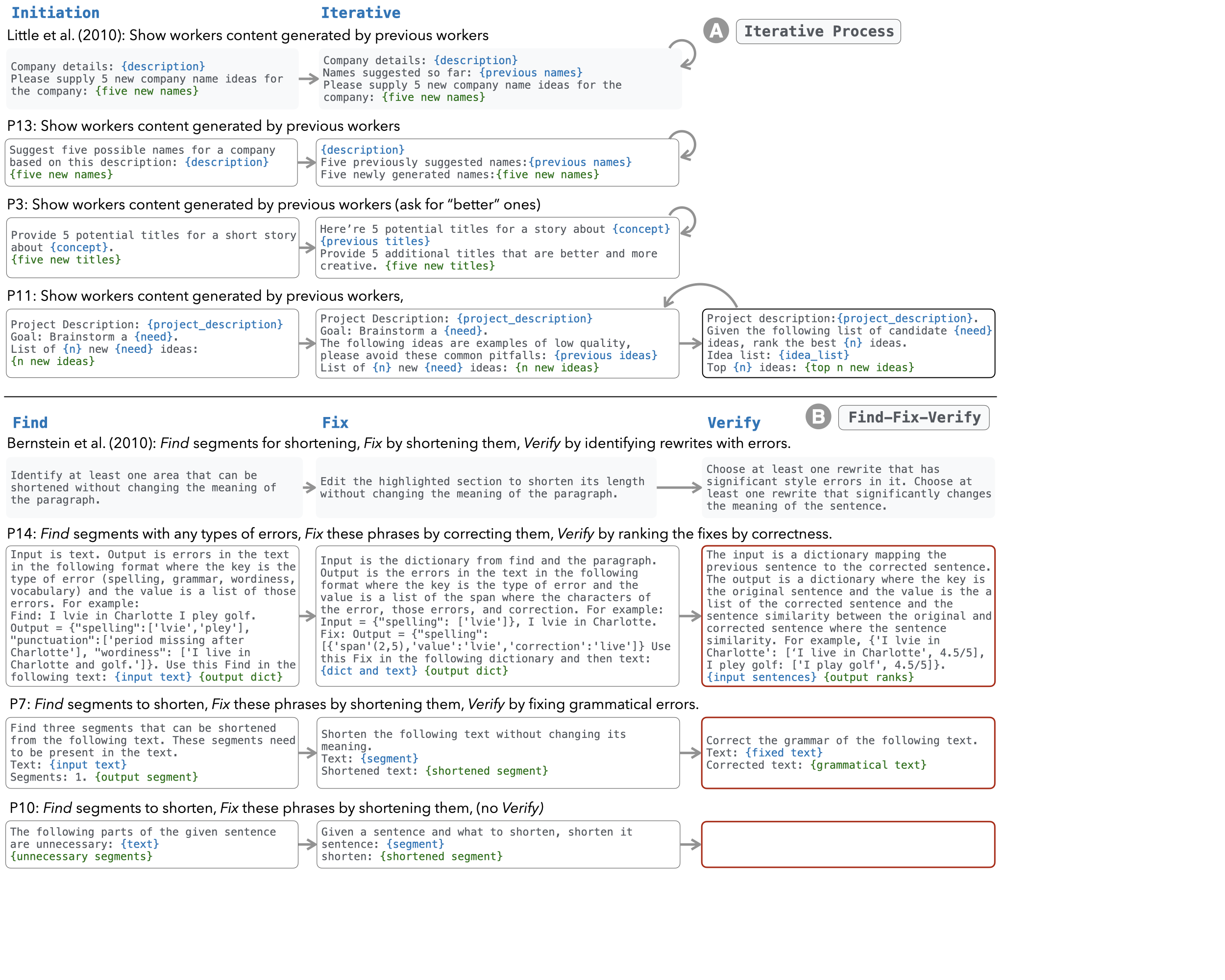}
\vspace{-5pt}
\caption{The original pipeline and the LLM replications for (A) \pIterateParallel~\cite{little2010exploring} and (B) \pf~\cite{bernstein2010soylent}. While only P11 diverged from the original \pIterateParallel by adding a condition about how previous results should be ranked and used in subsequent steps, students replicating \pf all had different Verify steps (marked in red box). The chains are slightly simplified for readability.}
\label{fig:prompt_variance}
\Description{A figure that highlights how crowdsourcing pipelines may be adapted in LLM chains replications.}
\end{figure*}

\subsection{Replication Overview: Partial Success}
\label{subsec:repl_overview}

As shown in Table~\ref{tab:pipelines}, all the pipelines are replicable with LLMs --- each pipeline has at least one correct and effective replication.
We illustrate this success with actual input-output sequences from student-preferred LLM chain replications enumerated in the table.
These findings re-iterate that LLMs can now perform tasks once thought exclusive to humans~\cite{bubeck2023sparks}.

Several students reported multiple chains they experimented with, echoing the prior observation that chains enable rapid prototyping~\cite{wu2022promptchainer}. 
Some of their explorations focused on single steps (\eg P7 in \pf choosing between different wordings among ``fragment'',  ``clauses'', ``substrings'' etc.), while others experimented with globally redesigning certain pipeline connections (\eg P11 in \pIterateParallel varied how the prior results should be passed onto the next step).
Interestingly, by examining students' final submissions and their own reflections, it becomes evident that (students believe) certain pipelines require adjustments (e.g., \pIterative, and \pf), while others can be replicated more literally (e.g., \pMapReduce).

That said, most pipelines did not achieve 100\% success or effectiveness. 
Students largely attributed the replication failure to prompting challenges --- \quoteinline{translating the pipeline into a LLM required a lot of work in terms of figuring out the correct prompt + the pre-processing that was required in order to move from one step in the pipeline to the next} (P14, \pf).
However, we believe there are more nuanced reasons underlying these prompting difficulties. 
In the following sections, we delve into several qualitative observations that have emerged from the replication practice and reflect on their implications (an overview of observations and opportunities are in Table~\ref{table:summary_findings}).

\subsection{Replication Variance: Impacted by students' perceptions on LLM capabilities}
\label{subsec:repl_variance}

Interestingly, we found some pipelines have more replication variance than others (i.e., different students replicated the same pipeline differently).
For instance, while both papers had adequate replication details, the three participants replicating \pIterateParallel arrived at similar chains. The only difference was created by P11 who added a step for choosing top previous results to show subsequent workers, without changing original steps.

In contrast, the three students replicating \pf implemented different versions (Figure~\ref{fig:prompt_variance}):
P14 mostly followed \citet{bernstein2010soylent}'s descriptions (e.g., having a voting mechanism in the \emph{Verify} step for reducing human errors), but extended the \emph{Find} step to include a lot more types of writing issues. They also designed their prompts \quoteinline{using data structures that are easily understandable by a computer versus natural language,} because the LLM \quoteinline{has a background with computer code.}
P7, on the other hand, only dedicated the \emph{Find} step to locate phrases that can be shortened, and instead implemented the \emph{Verify} step to fix grammatical errors that arose during the preceding shortening steps.
They explained that they consciously reshaped the design because they believed that \quoteinline{LLMs do not have these issues [of the high variance of human efforts and errors].}
However, this belief is arguably inaccurate. Just like human noise, the non-deterministic nature of LLMs can also lead to varying results if the same prompt is executed multiple times (similar to multiple workers completing the same sub-task).
In fact, prior work has applied a majority votes similar to \emph{Verification} in \pf for eliminating LLM noise~\cite{wang2022self, yao2023tree}, indicating that this step will still be useful for resolving the exact same issue: to remove problematic rewrites (now generated with LLMs).
P10 similarly removed the Verify step, possibly for similar rationales.

\paragraphBold{Reflection: Establish task-specific and pipeline-specific best practices for using LLMs.} 
The different implementations of crowdsourcing pipelines with LLMs reveal diverse student assumptions about model capabilities and necessary instructions.
Indeed, with the rapid advancement of LLMs and prompting techniques, it is challenging to keep up with LLMs' capabilities and limitations, as well as how they can be applied to specific use cases.
Instead of forming static and general mental models about LLMs, practitioners should \emph{dynamically adjust} their understanding based on the specific context. 
To achieve this, practitioners can adopt a mindset that views LLMs as ``Jack of all trades, master of none/few''~\cite{kocon2023chatgpt}.
They could start with broad, general prompts assuming LLMs' basic competency, and refine instructions based on prompt testing outcomes. 
This could evolve from high-level commands to more detailed instructions, addressing any identified deficiencies.
In the context of \pf, it might be sufficient to implement the \emph{Find} step with a high-level command like ``output any errors in the text,'' without specifying error types.
Then, if dedicated prompt testing~\cite{ribeiro_2023} reveals instances where the general prompt falls short, practitioners can adjust their prompts to incorporate more specific instructions, such as textual instructions on corner cases, or employ prompt ensembling techniques~\cite{pitis2023boosted}.

On the other hand, it appears that students have overlooked the fact that LLM performs probabilistic generation during their replication practice, despite being aware of this through their own experiences and course instructions.
It is intriguing to observe how the non-deterministic nature of LLM tends to be disregarded, particularly when used in a chaining context.
This oversight may stem from a trade-off between creating prototype chain structures and fine-tuning individual prompts for each sub-task~\cite{wu2022promptchainer}: 
LLM's non-determinism is typically presented using model confidence or the probability associated with the generated output, which may become a secondary consideration when students can only pass on a single output to the next sub-task.
To address this, introducing LLM non-determinism as ``noises exposed through voting of multiple LLM workers'' could allow for integration of disagreement mitigation techniques like adaptive decision-making on the number of votes/annotations needed~\cite{lin2014re, nie2020can}.

\subsection{Replication Effectiveness: Affected by LLM vs. Human Strengths}
\label{subsec:repl_effective}

\looseness=-1
So, what are the \emph{actual} strengths and weaknesses of LLMs, and how do they affect replicated LLM chains?
We delve into students' reflections on the implementation effectiveness. 
We find that, not too surprisingly, crowdsourcing pipelines proven effective might require some redesigning to accommodate the unique capabilities of LLMs, which \emph{still differ} from humans'. This observation aligns with discussions in prior work~\cite{webson2023language}; however, with the comprehensive exploration of the task space through replications, two significant patterns now become more apparent:




\paragraphBold{LLMs need explicit information foraging.}
Multiple crowdsourcing pipelines require \emph{implicit} information selection and integration.
For example, in \pMapReduce, workers performing the \emph{Reduce} step had to remove unnecessary information to make the final paragraph coherent.
Despite the necessity, few pipelines involve such explicit sub-tasks for selection.
This might be because humans are capable of implicit information filtering, re-ranking, and selection~\cite{pirolli1999information, sperber1986relevance, marchionini1995information}.
When it is clear that certain pieces are low-quality, out-of-place, or redundant, humans would proactively remove the unnecessary parts so as to retain a reasonable cognitive load.
In contrast, LLMs struggle with information foraging, and tend to constantly accumulate context and produce outputs with mixed quality.
Students observed these deficiencies at three levels and proposed possible changes:
\begin{itemize}[nosep, leftmargin=1.1em,labelwidth=*,align=left]

\item \emph{Fail to mitigate low-quality intermediate results. }
For example, when writing paragraphs with \pRecrusive, P4 found that even conflicting information from different sub-tasks would get integrated into the final writeup, resulting in incoherence (e.g., claiming the university mascot to be both a Scot and an owl).
Several students stressed the need for intermediate quality control, for \quoteinline{reducing the unpredictability of the model.} (P13, \pIterateParallel).

\item \emph{Fail to selectively perform a subset of sub-tasks.}
This is most visible in \pHumor, which, in its original design, required workers to \emph{self-select and sort} a subset of sub-tasks (eight in total) into an effective flow.
Among the four students replicating it, only P17 noticed that the sub-tasks have \quoteinline{no clear structure in the execution order of these micro tasks}, and successfully implemented a chain of four sub-tasks. Other students agreed that eight sub-tasks aggregated too much information, and P18 later reflected that \quoteinline{the steps should not be in such a strict order.}

\item \emph{Fail to meet multiple requirements in one sub-task.} 
Excessive requirements can collide. 
In the aforementioned \pHumor case, integrating results from too many sub-tasks may lead to certain options dominating others, e.g., the LLM can \quoteinline{focus on turning the joke into being sarcastic, which takes away the humor from the joke} (P5). 
Similarly, P14 (in \pf) implemented their \emph{Find} Step (Figure~\ref{fig:prompt_variance}) to simultaneous searching for multiple issues, which led the LLM to prioritize spelling errors and miss wordiness problems. 
Explicitly stating the top criteria seems important.

\end{itemize}

\paragraphBold{LLMs are more sensitive to comparison-based than humans.}
As commonly observed, LLMs are still sensitive to minor paraphrases (\eg P7 in \pf prototyped different wordings among ``fragment'',  ``clauses'', ``substrings'' etc. in their prompt).
However, on the flip side, LLMs are quite responsive to comparison-based instructions.
We will use \pIterateParallel for illustration.
In its original design, \citet{little2010exploring} reported anchoring bias to be an inherent limitation of the pipeline: ``perhaps owing to the fact that crowdworkers will iterate \& improve upon existing ideas, the variance is lower.'' 
All three students replicating this pipeline made similar observations but also found that such bias could be mitigated just with straightforward instructions.
For example, P11 initially observed that the pipeline \quoteinline{tends to converge on a specific theme,} but was able to redirect the model with a simple prompt: ``The following ideas are examples of low quality, please avoid these common pitfalls.''
Similarly, P3 was pleasantly surprised by how effective it is to simply \quoteinline{ask for outputs that differ from the initial set} --- \quoteinline{I was originally concerned that providing examples would `prime' the model to generate only examples in the same format, but it seems that this is not an issue in practice.}
Such simple instructions are unlikely to work for crowdworkers who are trapped by their personal biases~\cite{wu2021polyjuice}.

This sensitivity to adjectives such as ``different'' and ``diverse'' warrants further exploration. One peer grader highlighted this by suggesting, \quoteinline{If we're allowed to make suggestions, we could ask for titles that are happier, more obtuse, and funnier, which goes beyond traditional crowdsourcing methods.} 
This aligns with existing prompting techniques like Self-Refine~\cite{madaan2023self}, where LLMs critique their own outputs to generate improved versions focusing on specific dimensions.

\paragraphBold{Reflection: Examine effects of instruction tuning, and train humans for complementarity.} 
While differences between humans and LLMs are expected, it is interesting how some of these disparities arise from the goal of training LLMs to mimic human behavior.
For example, methods like Reinforcement Learning from Human Feedback~\cite[RLHF][]{ouyang2022training} use \emph{human preferences} to enhance LLMs' ability to follow instructions.
This might have simultaneously enabled LLMs to iterate on content based on abstract comparison commands \emph{more effectively than humans}, who often get trapped by cognitive bias or struggle with ambiguous or vague instructions~\cite{gershman2015computational}.
That said, it is unclear whether LLM generations are always \emph{better} in these cases, as these models are also biased by their training and can have polarized stands~\cite{jiang2022evaluating, santurkar2023whose}.

\looseness=-1
Branching out from this observation, it would be interesting to explore potential ``side-effects'' of the LLM training schema.
Prior work has highlighted the trade-off between few-shot vs. zero-shot capabilities and the need to train LLMs with multi-faceted human feedback~\cite{wu2023fine}. Considering LLMs' need for explicit information foraging, another worthy line of investigation would be the completeness and clarity of instructions.
As most existing instruction tuning datasets prioritize high-quality and precise instructions~\cite{longpre2023flan}, it remains unclear how LLMs would respond to ill-defined prompts or instructions containing irrelevant information.
It might be interesting to examine how LLMs can be trained using a ``chain-of-instruction-clarification'' approach, similar to the back-and-forth dialogues employed by humans to elicit design requirements. For instance, incorporating a sub-task that involves humans clarifying the top criteria could potentially enhance LLMs' ability to handle multiple requirements effectively.

The split of strengths also calls for \emph{human-LLM complementarity}. Instead of humans or LLMs completing all sub-tasks, an effective task delegation among a mixture of different ``workers'' might be useful.
For example, P15 in \pHumor noticed the partial effectiveness of their LLM chain: It excelled at \quoteinline{extracting relevant attributes of a news headline and brainstorming associated concepts} but failed at translating them into actual jokes.
As such, explicitly training humans to identify and develop skills complementary to LLM strengths could be an interesting direction to pursue~\cite{bansal2021does, ma2023ai,liu2023wants}.
Note that this complementarity can occur between humans and \emph{a variety of} LLMs.
For example, P3 in \pIterateParallel found that while using a weaker model either alone or in a pipeline resulted in poor performance, \quoteinline{when I provided examples from a stronger model as the previous examples [for the weaker model to iterate on], the performance dramatically improved.}
This shows even less capable models can be effective teammates if given the appropriate task --- ``All models are wrong, but some are useful.''~\cite{box1976science}.

\subsection{Replication Challenge: Multi-Modal Regulations vs. Textual Instructions}
\label{subsec:repl_challenge}

When reflecting on challenges in LLM replication, four students mentioned the difficulty of creating structured input/output formats. 
For example, P7 (replicating \pf) described including a constraint in their prompt: ``These segments need to be present in the text.''
They stressed its importance in the reflection: \quoteinline{Without this prompt, the returned segments are often sentences dramatically restructured based on the original text, making it difficult to insert them back into the original text after the fix step.}
Similarly, P6 in \pSRL said \quoteinline{the major weakness of these prompts was the challenge of extracting structured information out, especially for the pipeline models.}

\looseness=-1
It is worth considering why human workers, who are as (if not more) ``generative'' as LLMs, are capable of producing structured inputs and outputs. 
Essentially, all of the LLM replications of crowdsourcing pipelines are \emph{partial} --- 
the assignment focuses only on replicating the instructions of the crowdsourcing pipeline, while other components of crowdsourcing are disregarded.
Specifically, nearly all crowdsourcing pipelines inherently include constraints introduced by the user interface. 
For example, in \pf, the \emph{Find} step prompts crowdworkers to identify areas for abbreviation through \emph{mouse selection on text}, guaranteeing that the segment is precisely extracted from the original document. 
Similarly, \citet{he2015question} required annotators to label their questions and answers in a spreadsheet interface with limited answer length and predetermined question options.
These ensure that all the answers can be \emph{short phrases} to \emph{predictable questions}.
Meanwhile, since LLM modules/workers are solely driven by textual instructions, they need additional regulation to compensate for the absence of UI restrictions. 
Some students offered textual versions of syntactic constraints, e.g., \quoteinline{stricter templates (such as the use of a [MASK] token) would make crowdwork-style pipelines much easier.} (P11, \pIterateParallel).
Other ways might also be possible, e.g., transforming generative tasks into multiple-choice tasks so the LLM only outputs a single selection.

\paragraphBold{Reflection: Alignment in instruction modality, and its role in human simulation.} With the emergence of multi-modal models~\cite{openai2023gpt4, ramesh2022hierarchical}, it becomes crucial to not only contemplate the alignment between humans and models in terms of instruction following but also to explore the optimal instruction modality that aligns with human intuition.
For example, while LLMs have automated some interactions with visualization, prior work has found that users need mouse events to resolve vague references in their natural language commands (``make \emph{this bar} blue''~\cite{wang2022towards, kumar2017towards}).
Instead of converting such actions into textual instructions, it would be more advantageous to shift towards visual annotations.

Such challenges also have an impact on the practical applications of LLMs. 
In the ongoing discussions regarding whether LLMs can faithfully simulate humans, researchers are exploring LLMs as preliminary test subjects to refine study instructions and designs~\cite{hamalainen2023evaluating}.
Indeed, this direction is valuable---Just like in Figure~\ref{fig:prompt_variance}, both humans and LLMs need ``prompting'' to complete tasks.
Nevertheless, our findings indicate that such a transition may not be straightforward: 
Since LLMs only respond to textual instructions, an important post-processing step might be required to map LLM instructions into multi-modal constraints for humans.
For example, instruction ``extract exact sentences'' might need to be mapped to an interface design that involves selecting specific phrases, and ``paraphrase the main idea'' would require disabling copy-pasting from the text to discourage direct repetition and encourage users to provide their own input.
On one other hand, as mentioned before, LLMs and humans may respond differently to the same instructions, making them unreliable for precisely mimicking human reactions.
We suspect LLMs can be useful for helping study designers reflect on their \emph{high-level requirements} (e.g., determining what types of human responses to collect), but the literal instruction has to be redesigned.
Exploring which parts of the user study design can be prototyped using LLMs could be valuable.


\section{Discussion and Conclusion}

In this study, we explore using LLMs to replicate crowdsourcing pipelines in a course assignment. We find that while modern models can simulate human annotation to some extent in "human computation algorithms," their success varies widely based on task complexity. LLMs often perform unpredictably and struggle with multimodal cues crucial for accurate human annotation.

Our qualitative findings indicate two important points:
First, examining LLMs within established pipelines or workflows allows for a more straightforward understanding of their strengths and weaknesses, as different pipeline components have different requirements.
Second, when utilizing LLMs to simulate human computation, it is advantageous to not only focus on the inherent alignment between human and LLM outputs but also consider aligning additional scaffolds. 
This involves adapting existing techniques that tackle challenges such as misinterpretation of instructions by humans, noise in human responses, and the need to incorporate multi-modal constraints for humans.
Still, due to the setup of the course assignment, the LLM chain qualities varied greatly by students' efforts and expertise.
In addition, given the restricted sample size, quantitative analyses would have yielded limited significance.
Future research could systematically explore which crowdsourcing pipeline components benefit from LLM versus human involvements.

From an education perspective, we found having students interact with LLMs helped calibrate their confidence in these models---Many students conveyed their frustration when LLMs did not perform as reliably as they had anticipated. 
We hope the work can inspire future exploration on allowing students to interact with LLMs and gain awareness of these models' mistakes, thereby facilitating a constructive learning process and preventing excessive reliance on LLMs.

\begin{acks}
The work was supported by gift funds from the OpenAI Research Credit Program and Amazon Research. 
We thank the reviewers for their constructive feedback on the paper. 
\end{acks}

\bibliographystyle{ACM-Reference-Format}
\bibliography{ref}



\end{document}